\documentclass[11pt,a4paper]{article}
\usepackage[hyperref]{acl2019}
\usepackage{times}
\usepackage{latexsym}
\usepackage{amsmath}
\usepackage{amssymb}
\usepackage{mathrsfs}
\usepackage{url}
\usepackage{multirow}
\usepackage{varwidth}
\usepackage{color}
\usepackage{appendix}
\usepackage{enumitem}
\setitemize[1]{itemsep=0pt,partopsep=0pt,parsep=\parskip,topsep=0pt}

\usepackage{graphicx}
\graphicspath{ {./appendix/} }

\aclfinalcopy 

\newcommand*{\affaddr}[1]{#1} 

\newcommand*{\email}[1]{\texttt{#1}}

\newcommand*{\alter}[1]{#1}

\title{Proactive Human-Machine Conversation \\ with Explicit Conversation Goals}

\author{
Wenquan Wu, Zhen Guo, Xiangyang Zhou, Hua Wu, \\
\bf Xiyuan Zhang, Rongzhong Lian and Haifeng Wang \\
\affaddr{Baidu Inc., Beijing, China} \\
\email{$\left\{\begin{varwidth}{12cm}\centering wuwenquan01,guozhenguozhen,zhouxiangyang,wu\_hua\end{varwidth}\right\}$@baidu.com}\\
\email{$\left\{\begin{varwidth}{12cm}\centering zhangxiyuan01,lianrongzhong,wanghaifeng\end{varwidth}\right\}$@baidu.com}
}

\date{}

\begin{document}

\maketitle

\begin{abstract}

Though great progress has been made for human-machine conversation, current dialogue system is still in its infancy: it usually converses passively and utters words more as a matter of response, rather than on its own initiatives.
In this paper, we take a radical step towards building a human-like conversational agent: endowing it with the ability of proactively leading the conversation (introducing a new topic or maintaining the current topic).
To facilitate the development of such conversation systems, we create a new dataset named DuConv where one acts as a conversation leader and the other acts as the follower. 
The leader is provided with a knowledge graph and asked to sequentially change the discussion topics, following the given conversation goal, and meanwhile keep the dialogue as natural and engaging as possible.
DuConv enables a very challenging task as the model needs to both understand dialogue and plan over the given knowledge graph.
We establish baseline results on this dataset (about 270K utterances and 30k dialogues) using several state-of-the-art models. Experimental results show that dialogue models that plan over the knowledge graph can make full use of related knowledge to generate more diverse multi-turn conversations.
The baseline systems along with the dataset are publicly available \footnote{ https://github.com/PaddlePaddle/models/tree/develop/\\
PaddleNLP/Research/ACL2019-DuConv}.

\end{abstract}

\section{Introduction}

Building a human-like conversational agent is one of long-cherished goals in Artificial Intelligence (AI) \cite{turing2009computing}. Typical conversations involve exchanging information \cite{zhang2018personalizing}, recommending something \cite{li2018towards}, and completing tasks \cite{bordes2016learning}, most of which rely on background knowledge. However, many dialogue systems only rely on utterances and responses as training data, without explicitly exploiting knowledge associated with them, which sometimes results in uninformative and inappropriate responses \cite{wang2018learning}. Although there exist some work that use external background knowledge to generate more informative responses {\cite{P18-1138,journals/corr/YinJLSLL15,DBLP:journals/corr/abs-1709-04264}, these systems usually generate responses to answer questions instead of asking questions or leading the conversation.  In order to solve the above problems, some new datasets have been created, where external background knowledge is explicitly linked to utterances \cite{dinan2018wizard,moghe2018towards}, to facilitate the development of knowledge aware conversation models. With these datasets, conversation systems can be built to talk with humans given a topic based on the provided external knowledge. Unlike task-oriented systems \cite{bordes2016learning}, these conversation systems don't have an explicit goal to achieve, thereof not able to plan over the background knowledge.

\begin{figure*}[h]
\centering
\includegraphics[width=0.95\textwidth,height=0.2\textheight]{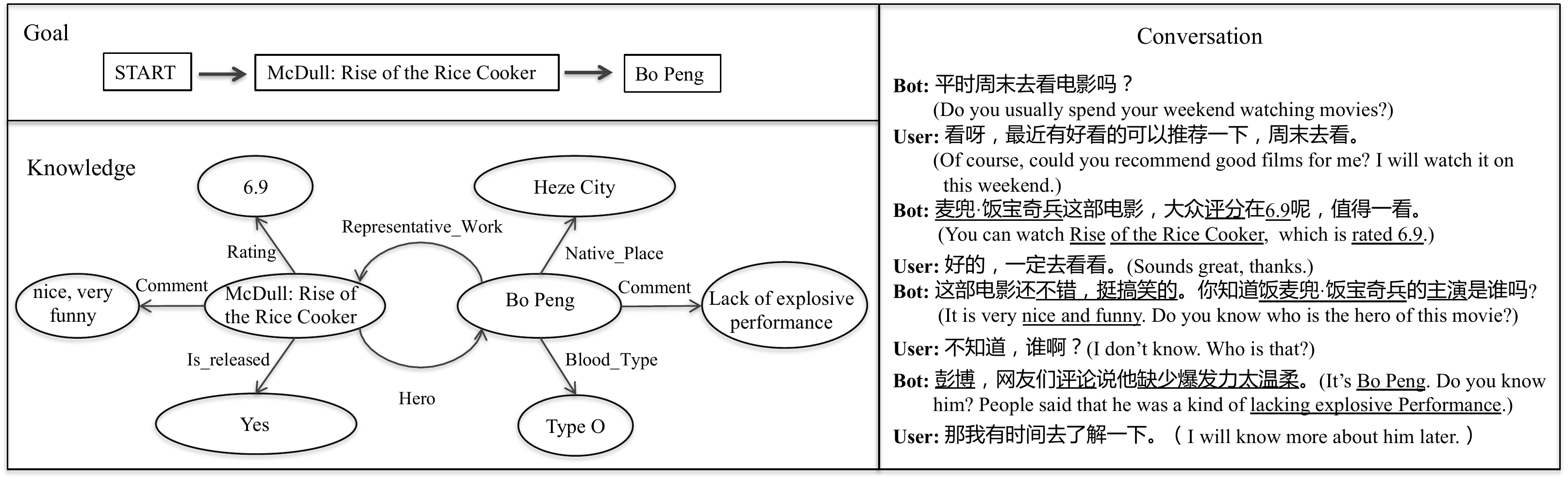}
\caption{One conversation generated by two annotators, one of which was given a goal and related knowledge.}
\label{fig:1}
\end{figure*}

In this paper, we take a radical step towards building another type of human-like conversational agent: endowing it with the ability of proactively leading the conversation with an explicit conversation goal. 
To this end, we investigate learning a proactive dialogue system by planning dialogue strategy over a knowledge graph.  Our assumption is that reasoning and planning with knowledge are the keystones to achieve proactive conversation. For example, when humans talk about movies, if one person learns more about some movies, he/she usually leads the conversation based on one or more entities in the background knowledge and smoothly changes the topics from one entity to another. In this paper, we mimic this process by setting an explicit goal as a knowledge path ``\emph{[start] $\to$ topic\_a $\to$ topic\_b}", which means that one person leads the conversation from any starting point to \emph{topic\_a} and then to \emph{topic\_b}. Here \emph{topic} represents one entity in the knowledge graph.

With this in mind, we first build a knowledge graph which combines factoid knowledge and non-factoid knowledge such as comments and synopsis about movies. To construct the knowledge graph, we take a factoid knowledge graph (KG) as its backbone and align unstructured sentences from the non-factoid knowledge with entities. Then we use this KG to facilitate knowledge path planning and response generation, as shown in Figure 1. Based on this knowledge graph, we create a new knowledge-driven conversation dataset, namely the Baidu Conversation Corpus (\textbf{DuConv}) to facilitate the development of proactive conversation models.  Specifically, DuConv has around 30k multi-turn conversations and each dialog in the DuConv is created by two crowdsourced workers, where one plays the role of the conversation leader and the other one acts as the conversation follower.
At the beginning of each conversation, the leading player is assigned with an explicit goal, i.e., to sequentially change the conversation topic from one to another, meanwhile keeping the conversation as natural and engaging as possible. 
The conversation goal is a knowledge path comprised of two topics and structured as ``\emph{[start] $\to$ topic\_a $\to$ topic\_b}" and the leading player is also provided with related knowledge of these two topics.
For each turn in the conversation, the leading player needs to exploit the provided knowledge triplets to plan his/her conversation strategy and construct responses to get closer to the target topic, while the follower only needs to respond according to the contexts without knowing the goal. 

Figure ~\ref{fig:1} illustrates one example dialog in DuConv. It can be seen that DuConv provides a very challenging task: the conversational agents have to fully exploit the provided knowledge to achieve the given goal. To test the usability of DuConv, we propose a knowledge-aware neural dialogue generator and a knowledge-aware retrieval-based dialogue system, and investigate their effectiveness. Experimental results demonstrate that our proposed methods can proactively lead the conversation to complete the goal and make more use of the provided knowledge. 

To the best of our knowledge, it is the first work that defines an explicit goal over the knowledge graph to guide the conversation process, making the following contributions:
\begin{itemize}
\setlength{\itemsep}{0pt}
\setlength{\parsep}{0pt}
\setlength{\parskip}{0pt}
	\item A new task is proposed to mimic the action of humans that lead conversations over a knowledge graph combining factoid and non-factoid knowledge, which has a wide application in real-world but is not well studied.
	\item A new large-scale dataset named DuConv is constructed and released to facilitate the development of knowledge-driven proactive dialogue systems.
	\item We propose knowledge-aware proactive dialogue models and conduct detailed analysis over the datasets. Experimental results demonstrate that our proposed methods make full use of related knowledge to generate more diverse conversations.
\end{itemize}

\section{Related Work}

Our related work is in line with two major research topics,  \emph{Proactive Conversation} and \emph{Knowledge Grounded Conversation}.

\subsection{Proactive Conversation}

The goal of proactive conversation is endowing dialogue systems with the ability of leading the conversation. Existing work on proactive conversation is usually limited to specific dialogue scenarios.  Young et al. \shortcite{young2013pomdp}, Mo et al. \shortcite{bordes2016learning} and Bordes et al. \shortcite{mo2018personalizing} proposed to complete tasks more actively, like restaurant booking, by actively questioning/clarifying the missing/ambiguous slots. Besides the task-oriented dialogue systems, researchers have also investigated building proactive social bots to make the interaction more engaging. Wang et al., \shortcite{wang2018learning} explored to ask good questions in open-domain conversational systems. Li et al., \shortcite{li2018towards} enabled chatbots to recommend films during chitchatting. Unlike the existing work, we proposed to actively lead the conversation by planning over a knowledge graph with an explicit goal. We also create a new dataset to facilitate the development of such conversation systems.

\subsection{Knowledge Grounded Conversation}

Leveraging knowledge for better dialogue modeling has drawn lots of research interests in past years and researchers have shown the multi-fold benefits of exploiting knowledge in dialogue modeling. One major research line is using knowledge to generate engaging, meaningful or personalized responses in chitchatting \cite{ghazvininejad2018knowledge,vougiouklis2016neural,zhou2018commonsense,zhang2018personalizing}. In addition to proposing better conversation models, researchers also released several knowledge grounded datasets \cite{dinan2018wizard,moghe2018towards}. Our work is most related to Mogh et al., \shortcite{moghe2018towards} and Dinan et al., \shortcite{dinan2018wizard}, where each utterance in their released datasets is aligned to the related knowledge, including both structured triplets and unstructured sentences. We extend their work, by including the whole knowledge graph into dialogue modeling and propose a new task of proactively leading the conversation via planning over the knowledge graph in this paper.

\section{DuConv}

\begin{table}[]
\centering
\begin{tabular}{|c|c|}
\hline 
\# dialogs & 29858 \\ \hline
\# utterances & 270399 \\ \hline
average \# utterances per dialog & 9.1 \\ \hline
average \# words per utterance & 10.6 \\ \hline
average \# words per dialog & 96.2 \\ \hline
average \# knowledge per dialogue & 17.1 \\
\hline
\end{tabular}
\caption{Overview of the conversation dataset DuConv.}
\label{table:DuConv}
\end{table}

In this section, we describe the creation of DuConv in details.  It contains four steps: knowledge crawling, knowledge graph construction, conversation goal assignment, and conversation crowdsourcing. We limit the dialogue topics in DuConv to movies and film stars, and crawl this related knowledge from the internet. Then we build our knowledge graph with these crawled data. After constructing our knowledge graph, we randomly sample two linked entities to construct the conversation goal, denoted as ``\emph{[start] $\to$ topic\_a $\to$ topic\_b}", and ask two annotators to conduct knowledge-driven conversations, with one playing as the conversation leader and the other one playing as the follower. The leader needs to change the conversation topics following the conversation goal and meanwhile keep the conversation as engaging as possible. All those conversations are recorded and around 30k conversations are finally used in DuConv after filtering dirty/offensive parts. Table ~\ref{table:DuConv} summarizes the main information about DuConv.

\subsection{Knowledge Crawling}

We crawled the related knowledge information from the website MTime.com\footnote{http://www.mtime.com/}, which records the information of most films, heroes, and heroines in China. We collect both structured knowledge (such as ``\emph{Harry Potter}" is ``\emph{directed\_by}" ``\emph{Chris Columbus}") as well as unstructured knowledge including short \emph{comments} and \emph{synopsis}. We filter out the dirty or offensive information and further normalize some of the numbers (such as the values of \emph{rating}) into discrete symbols (\emph{good, fair, bad}) to facilitate the use of this kind of knowledge. In summary, we crawl more than 91k films and 51k film stars, resulting in about 3.6 million knowledge triplets, the accuracy of which is over 97\% \footnote{We randomly sampled 100 triplets and manually evaluated them.}. 

\subsection{Knowledge Graph Construction}

\begin{table}[]
\centering
\begin{tabular}{|c|c|}
\hline
\# entities & 143627 \\ \hline
\# movies & 91874 \\ \hline
\# person names  & 51753 \\ \hline
\# properties & 45 \\ \hline
\# spo & 3598246 \\ \hline
average \# spo per entity & 25 \\
\hline
\end{tabular}
\caption{Overview of the knowledge graph in DuConv.}
\label{table:knowledge}
\end{table}

After the raw data collection, we construct a knowledge graph. Our knowledge graph is comprised of multiple SPO (\emph{Subject, Predicate, Object}) knowledge triplets, where objects can be factoid facts and non-factoid sentences such as comments and synopsis. The knowledge triplets in our graph can be classified into:
\begin{enumerate}
	\item \textbf{Direct triplets}: widely-used knowledge triplets, such as (``Harry Potter and the Sorcerer Stone", "directed\_by", "Chris Columbus"), akin to most existing knowledge graphs, with the exception that the objects can be sentences such as short \emph{comments} and \emph{synopsis}.
	\item \textbf{Associated triplets}: if two entities share the same predicate and the same object in their triplets, then we create a virtual triplet like ("Harry Potter and the Sorcerer Stone", "directed\_by\_Chris Columbus", "Home Alone") by combining the two original triplets.
\end{enumerate}
We call the direct triplets as \emph{one-step relation} and associated triplets as \emph{two-step relation}. Table ~\ref{table:knowledge} lists the main information of our knowledge graph. 

\subsection{Conversation Goal Assignment}

\begin{figure*}[h]
\centering
\includegraphics[width=\textwidth,height=0.25\textheight]{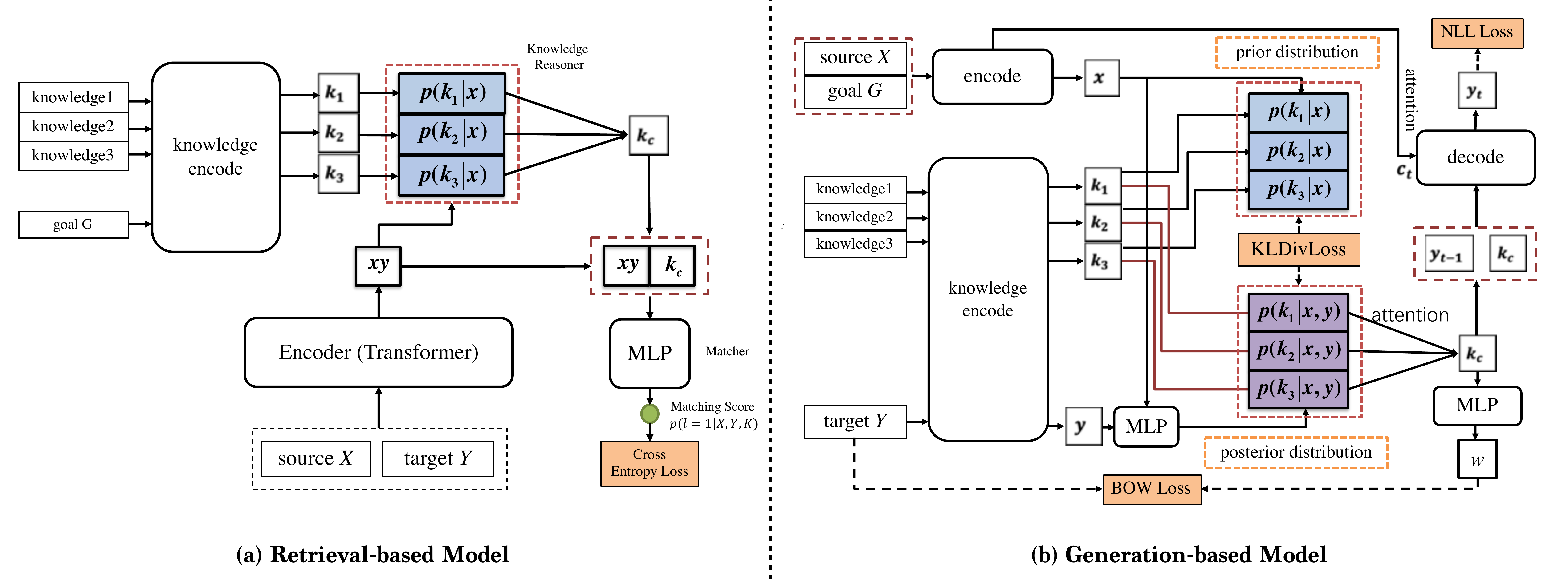}
\caption{The retrieval-based model and generation-based model.}
\label{fig:3}
\end{figure*}

Given the knowledge graph, we sample some knowledge paths, which are used as conversation goals. Specifically, we focus on the simple but challenging scenario: naturally shifting the topics twice, i.e., from ``\emph{[start]}" state to ``\emph{topic\_a}" then finally to ``\emph{topic\_b}". We sample two linked entities in our knowledge graph as `\emph{topic\_a}" and ``\emph{topic\_b}" to construct the knowledge path. About 30k different knowledge paths are sampled and used as conversation goals for knowledge-driven conversation crowdsourcing, where half of the knowledge paths are from the one-step relation set while the other half are from the two-step relation set.

\subsection{Crowdsourcing}

Unlike using self-play in dataset construction \cite{ghazvininejad2018knowledge}, we collect lots of \alter{crowdsourced workers} to generate the dialogues in DuConv \footnote{The workers are collected from a Chinese crowdsourcing platform http://test.baidu.com/. The workers are paid 2.5 Chinese Yuan per conversation.}. For each given conversation goal, we assign two \alter{workers} different roles: 1) the conversation leader and 2) the follower. The leader is provided with the conversation goal and its related background knowledge in our knowledge graph, and then asked to naturally shift the conversation topic following the given conversation goal. The follower is provided with nothing but the dialogue history and only has to respond to the leader. The dialogue will not stop until the leader achieves the conversation goal. We record conversation utterances together with the related knowledge triplets and the knowledge path, to construct the whole dataset of DuConv. 

\section{Methods}

To enable neural dialogue systems to converse with external background knowledge, we propose two models: a retrieval-based model and a generation-based model, by introducing an external memory module for storing all related knowledge, making the models select appropriate knowledge to enable proactive conversations. 
Figure ~\ref{fig:3} shows the architectures of our proposed knowledge-aware response ranking model as well as our response generation model. We will give a detailed description of those two knowledge-aware models in next two sub-sections.

\subsection{Retrieval-based Model}

Given a dialogue context $X$, the retrieval-based dialogue system responds to that context via searching for the best response $Y$ from DuConv. Thus retrieval-based dialogue system often has a pipeline structure with two major steps: 1) retrieve response candidates from a database and 2) select the best one from the response candidates \cite{zhou2018multi}. In our retrieval-based method, the candidate responses are collected similar to most existing work \cite{wu2017sequential,zhou2018multi} with one notable difference that we normalize the entities with their entity types in the knowledge graph to improve generalization capabilities.

For each retrieved candidate response $Y$, the goal of our response ranker is to measure if $Y$ is a good response to the context $X$ considering the given dialogue goal $G=[start, topic\_a, topic\_b]$ and related knowledge $K$. The matching score measured by our knowledge-aware response ranker is defined as $p(l=1|Y, X, K, G)$. As shown in Figure ~\ref{fig:3}(a),  our knowledge-aware response ranker consists of four major parts, i.e., the context-response representation module (Encoder), the knowledge representation module (Knowledge Encoder), the knowledge reasoning module (Knowledge Reasoner) as well as the matching module (Matcher). 

The Encoder module has the same architecture as BERT  \cite{devlin2018bert}, it takes the context $X$ and candidate response $Y$ as segment\_a and segment\_b in BERT, and leverages a stacked self-attention to produce the joint representation of $X$ and $Y$, denoted as $xy$. Each related knowledge $knowledge_i$ is also encoded as vector representations in the Knowledge Encoder module using a bi-directional GRU \cite{chung2014empirical}, which can be formulated as $k_i = [\overrightarrow{h_T};\overleftarrow{h_0}]$, where $T$ denotes the length of knowledge, $\overrightarrow{h_T}$ and $\overleftarrow{h_0}$ represent the last and initial hidden states of the two directional GRU respectively. The dialogue goal is also combined with the related knowledge in order to fuse that information into response ranking.

To jointly consider context, dialogue goal and knowledge in response ranking, we make the context-response representation $xy$ attended to all knowledge vectors ${k_i}$ and get the attention distribution. For simplicity, the dialogue goal was treated as part of the knowledge used in the conversation.
\begin{align}
\small
\alter{
p(k_i|x,y) = \frac{exp(xy \cdot k_i )}{\sum_j exp(xy \cdot k_j)}
} 
\end{align}
\noindent
and fuse all related knowledge information into a single vector\alter{ $k_c = \sum_i p(k_i|x,y) * k_i$.} We view $k_c$ and $xy$ as the information from knowledge side and dialogue side respectively, and fuse those two kinds of information into a single vector via concatenation, then finally calculate the matching probability as:
\begin{equation}
\small
p(l=1|X,Y,K,G) = sigmoid(\textbf{MLP}([xy;k_c]))
\end{equation}
\noindent
Our knowledge-aware response ranker differs from most existing work in jointly considering the previous dialogue context, the dialogue goal as well as the related knowledge, which enables our model to better exploit knowledge to achieve the conversation goal. 

\subsection{Generation-based Model}

To generate a knowledge-driven dialogue response, we enhance the vanilla seq2seq model with an extra knowledge selection paradigm, Figure ~\ref{fig:3}(b) demonstrates the structure of our knowledge-aware generator, which is comprised of four parts: the \emph{Utterance Encoder}, the \emph{Knowledge Encoder}, the \emph{Knowledge Manager} and the \emph{Decoder}. 

For each given dialogue context $X$, along with the dialogue goal $G$ and related knowledge $K$, our knowledge-aware generator first encodes all input information as vectors in the Utterance Encoder and Knowledge Encoder. The encoding method in those two modules also uses bi-directional GRUs, akin to that in the retrieval-based method. Especially, the dialogue context $X$ and dialogue goal $G$ are fused into the same vector $x$ by sequentially concatenate $G$ and $X$ into a single sentence, then feed to the encoder. 

After encoding, our knowledge-aware generator starts to plan its dialogue strategy by considering which knowledge would be appropriate next. Practically, the generator can also conduct knowledge selection via attention mechanism as in the retrieval-based method. However, to force the model to mimic human in knowledge selection, we introduce two different distributions: 1) the \emph{prior distribution} $p(k_i | x)$ and the \emph{posterior distribution} $p(k_i | x, y)$. 
We take the prior distribution $p(k_i | x)$ as the knowledge reasoned by machines and the posterior distribution $p(k_i | x, y)$ as the knowledge reasoned by humans, and then force the machine to mimic human by minimizing the KLDivLoss between those two distributions, which can be formulated as:
\begin{align} 
\small
& p(k_i | x, y) = \frac{exp(k_i \cdot MLP([x;y]))}{\sum_{j=1}^N exp( k_j \cdot MLP([x;y]))}
\\
& p(k_i | x) = \frac{exp(k_i \cdot x)}{\sum_{j=1}^N exp(k_j \cdot x)}
\\
& \alter{L_{KL}(\theta) = \frac{1}{N} \sum_{i=1}^N p(k_i | x,y) log \frac{p(k_i | x, y)}{p(k_i | x)}}
\end{align}
\noindent
Given that knowledge distribution $p(k_i | x)$ and  $p(k_i | x, y)$, we fused all related knowledge information into a vector \alter{$k_c=\sum_i p(k_i|x,y)*k_i$ }, same as our retrieval-based method, and feed it to the decoder for response generation. In the testing phase, the fused knowledge is estimated by the formula  $k_c=\sum_i p(k_i|x)*k_i$ without gold responses . The decoder is implemented with the \emph{Hierarchical Gated Fusion Unit} described in \cite{yao2017towards}, which is a standard GRU based decoder enhanced with external knowledge gates. Besides the KLDivLoss, our knowledge-aware generator introduces two additional loss functions:
\begin{description}
\item \textbf{NLL Loss}: the Negative Log-Likelihood (NLL) $L_{NLL}(\theta)$ measures the difference between the true response and the response generated by our model.
\item \textbf{BOW Loss}: We use the BOW loss proposed by Zhao et al., \shortcite{zhao2017learning}, to ensure the accuracy of the fused knowledge $k_c$ by enforcing the relevancy between the knowledge and the true response. Specifically, let $w = \textbf{MLP}(k_c) \in \mathcal{R}^{|V|}$, where $|V|$ is the vocabulary size and we define:
\begin{equation} 
 p(y_t|k_c) = \frac{exp(w_{y_t})}{\sum_v exp(w_v) }
 \end{equation}
 
 \noindent
 Then, the BOW loss is defined to minimize:
\begin{equation} 
L_{BOW}(\theta) = - \frac{1}{m} \sum_{t=1}^m log p(y_t | k_c)
\end{equation}
\end{description}
\noindent
In summary, the final loss of our generative model is:
\begin{align}
\small
L(\theta) = L_{KL}(\theta)  + L_{NLL}(\theta)  + L_{BOW}(\theta) 
\end{align}

\begin{table*}
\centering
\small
\begin{tabular}{ |c|c|c|c|c|c|c| } 
 \hline
Methods &  \emph{Hits{@}1}  & \emph{Hits{@}3}  & PPL & F1/BLEU1/BLEU2 & DISTINCT 1\&2 & knowledge R/P/F1 \\ \hline
 \hline

retrieval w/o klg. & 45.84\% & 72.86\% & - &  33.08 / 0.280 / 0.147 & 0.121 / 0.376 & 8.69 / 39.30 / 13.73  \\ \hline
retrieval w/ klg. & 46.74\% & 75.32\% & - & 33.12 / 0.282 / 0.146& \textbf{0.122} / \textbf{0.388} & 8.54 / 37.93 / 13.47 \\ \hline
norm retrieval  & \textbf{50.92}\% & \textbf{79.02}\% & - & 34.73 / 0.291 / 0.156 & 0.118 / 0.373 & 8.85/ 38.00 / 13.88 \\ \hline
S2S w/o klg. & 24.88\% & 49.64\% & 20.16 & 26.43 / 0.187 / 0.100 & 0.032 / 0.088 & 4.59 / 30.00 / 7.73 \\ \hline
S2S w/ klg. & 30.58\% & 57.52\% & 13.53 & 32.19 / 0.226 / 0.140 & 0.064 / 0.168 & 5.89 / 36.31 / 9.85 \\ \hline
norm S2S & 31.26\% & 55.12\% & \textbf{10.96} & 39.94 / 0.283 / 0.186 & 0.093 / 0.222 & 7.52 / \textbf{42.74} / 12.34 \\ \hline
generation w/o klg. & 25.52\% & 50.14\% & 20.3 & 28.52 / 0.29 / 0.154 & 0.032 / 0.075 & 6.18 / 27.48 / 9.86 \\ \hline
generation w/ klg. & 31.90\% & 58.44\% & 27.3 & 36.21 / 0.32 / 0.169 & 0.049 / 0.144 & 8.67 / 35.90 / 13.62 \\ \hline
norm generation & 32.50\% & 58.50\% & 24.3 & \textbf{41.84} / \textbf{0.347} / \textbf{0.198} & 0.057 / 0.155 & \textbf{9.78} / 38.02 / \textbf{15.27} \\ \hline

\end{tabular}
\caption{Automatic evaluation results. klg. and norm stands for \emph{knowledge} and \emph{normalized} here. S2S stands for the vanilla sequence-to-sequence model.}
\label{table:results}
\end{table*}

\begin{table*}
\centering
\small
\begin{tabular}{ |c|c|c|c|c|c|c|} 
 \hline
 methods & \multicolumn{4}{|c|}{turn-level human evaluation} & \multicolumn{2}{|c|}{dialogue-level human evaluation} \\ \hline
 metrics & fluency  & coherence & informativeness & proactivity & goal complete & coherence \\ 
 scores & (0,1,2) & (0,1,2) & (0,1,2) &  (-1,0,1) & (0,1,2) &  (0,1,2,3)  \\ \hline
 \hline
norm retrieval & 1.93 & 1.41 & 0.86 & 0.80 & 0.90 & 1.92 \\ \hline
norm generation (s2s) & \textbf{2.00} & \textbf{1.89} & 0.74 & 0.86 & 1.14 & 2.01 \\ \hline
norm generation & 1.87 & 1.61 & \textbf{1.10} & \textbf{0.87} & \textbf{1.22} & \textbf{2.03} \\ \hline
\end{tabular}
\caption{ Turn-level and dialogue-level human evaluation results}
\label{table:human-results}
\end{table*}

\section{Experiments}

\subsection{Setting}

Our proposed models are tested under two settings: 1) automatic evaluation and 2) human evaluation. For automatic evaluation, we leverage several common metrics including BLEU, PPL, F1, DISTINCT1/2 to automatically measure the fluency, relevance, diversity etc. In our setting, we ask each model to select the best response from 10 candidates, same as previous work \cite{zhang2018personalizing}. Those 10 candidate responses are comprised of one true response generated by human-beings and nine randomly sampled ones from the training corpus. We measure the performance of all models using \emph{Hits{@}1} and  \emph{Hits{@}3}, same as Zhang et al., \shortcite{zhang2018personalizing}. Furthermore, we also evaluate the ability of exploiting knowledge of each model by calculating knowledge precision/recall/F1 scores.

The human evaluation is conducted at two levels, i.e., the turn-level human evaluation and the dialogue-level human evaluation. The turn-level human evaluation is similar to automatic evaluation. Given the dialogue context, the dialogue goal as well as the related knowledge, we require each model to produce a response according to the dialogue context. The responses are evaluated by three annotators in terms of fluency, coherence, informativeness, and proactivity. The coherence measures the relevance of the response and the proactivity measures if the model can successfully introduce new topics without destructing the fluency and coherence.

The dialogue-level evaluation is much more challenging.  Given a conversation goal and the related knowledge, each model is required to talk with a volunteer and lead the conversation to achieve the goal. For each model, 100 dialogues are generated.  The generated conversations are then evaluated by three persons in terms of two aspects: goal completion and coherence. The goal completion measures how good the conversation goal is achieved and the coherence scores the fluency of the whole dialogue.

All human evaluation metrics, except the turn-level proactivity and the dialogue-level coherence, has three grades: good(2), fair(1), bad(0).  For goal completion, ``2" means that the goal is achieved with full use of knowledge, ``1" means the goal is achieved by making minor use of knowledge, and ``0" means that the goal is not achieved. We additionally set the perfect grade (3) for dialogue-level coherence, to encourage consistent and informative dialogues. For proactivity, we also have three grades: ``1" means good proactivity that new topics related to context are introduced, ``-1" means bad proactivity that new topics are introduced but irrelevant to context, and ``0" means that no new topics are introduced. The detailed description of the human evaluation metrics can be found in the appendices.

\subsection{Comparison Models}

The compared models contain the vanilla seq2seq model, our proposed retrieval-based model as well as our proposed generation-based model\footnote{We also compared MemNet \cite{ghazvininejad2018knowledge}, whose performance is similar to Seq2Seq with knowledge. We omit it for space limit in this paper.}. Moreover, we normalize the train/valid/test data by replacing the specific two topics in the knowledge path with ``\emph{topic\_a}" and ``\emph{topic\_b}" respectively. Models using such normalized corpora are named as normalized models. To test the effectiveness of knowledge, we set up one ablation experiment, which removes all the knowledge triplets by replacing them with ``{UNK, UNK, UNK}".

\begin{table*}
\centering
\begin{tabular}{ |c|c|c|c|c|} 
 \hline
 
 \multicolumn{2}{|c|}{distribution statistics} & norm generation& norm seq2seq& norm retrieval\\ \hline
 \hline
 \multirow{3}{*}{goal completion} & 0 & 21\% & 14\% & \textbf{25\%} \\
 & 1 & 35\% & 26\% & \textbf{59\%} \\
 & 2 & \textbf{43\%} & 29\% & 15\% \\ \hline
 \multirow{2}{*}{knowledge used} & \# triplets & \textbf{2.46} & 1.51 & 2.28 \\
 & \# properties & \textbf{27} & 20 & 25 \\
\hline
\end{tabular}
\caption{Analysis on goal completion and knowledge exploition.}
\label{table:analysis-1}
\end{table*}

\subsection{Model Training}

All models are implemented using PaddlePaddle \footnote{It is an open source deep learning platform (https://paddlepaddle.org) developed by Baidu. Our code and data are available at https://github.com/PaddlePaddle/models/\\tree/develop/PaddleNLP/Research/ACL2019-DuConv.} and pytorch \cite{paszke2017automatic}, trained on a single GPU of NVIDIA Tesla K40. We set the vocabulary size to 30k for both retrieval-based and generation based methods. All hidden sizes, as well as embedding size, are set to 300, and the word embedding layer is initialized via word2vec\footnote{https://radimrehurek.com/gensim/models/word2vec.html} trained on a very large corpus. We apply Adam optimize for model training and the beam size for generative models are set to 10 during decoding.

\subsection{Results}

Table ~\ref{table:results} and Table ~\ref{table:human-results} summarize the experimental results on automatic evaluation and human evaluation. For human evaluation, we only evaluate the normalized models since they achieved better performances on our dataset. All human evaluations are conducted by three persons, where the agreement ratio (Fleiss' kappa \cite{fleiss1971measuring}) is from 0.37 to 0.86, with the lowest agreement on multi-turn coherence and others all above 0.6.  More details of these measures are available in the Appendix.

It can be seen that the retrieval-based model and the generation-based model have significantly different performances in terms of automatic evaluation and human evaluations. Retrieval-based model works better on Hits{@}K, however worse on F1 and BLEU compared to the generation-based model. This is perhaps caused by that fact that they are optimized on different metrics.  For human evaluation, it can be observed that the retrieval-based method is apparently worse than generation-based models. This is because the retrieved candidates limit the potential of the retrieval-based model. We also found that the methods using knowledge outperform those without using knowledge, which confirms the benefits of using background knowledge. It is very interesting that normalizing the ``\emph{topic\_a}" and ``\emph{topic\_b}" can significantly improve the performance for all models because of their generalization capability over the knowledge. 

\begin{figure*}[h]
\centering
\includegraphics[width=0.98\textwidth,height=0.3\textheight]{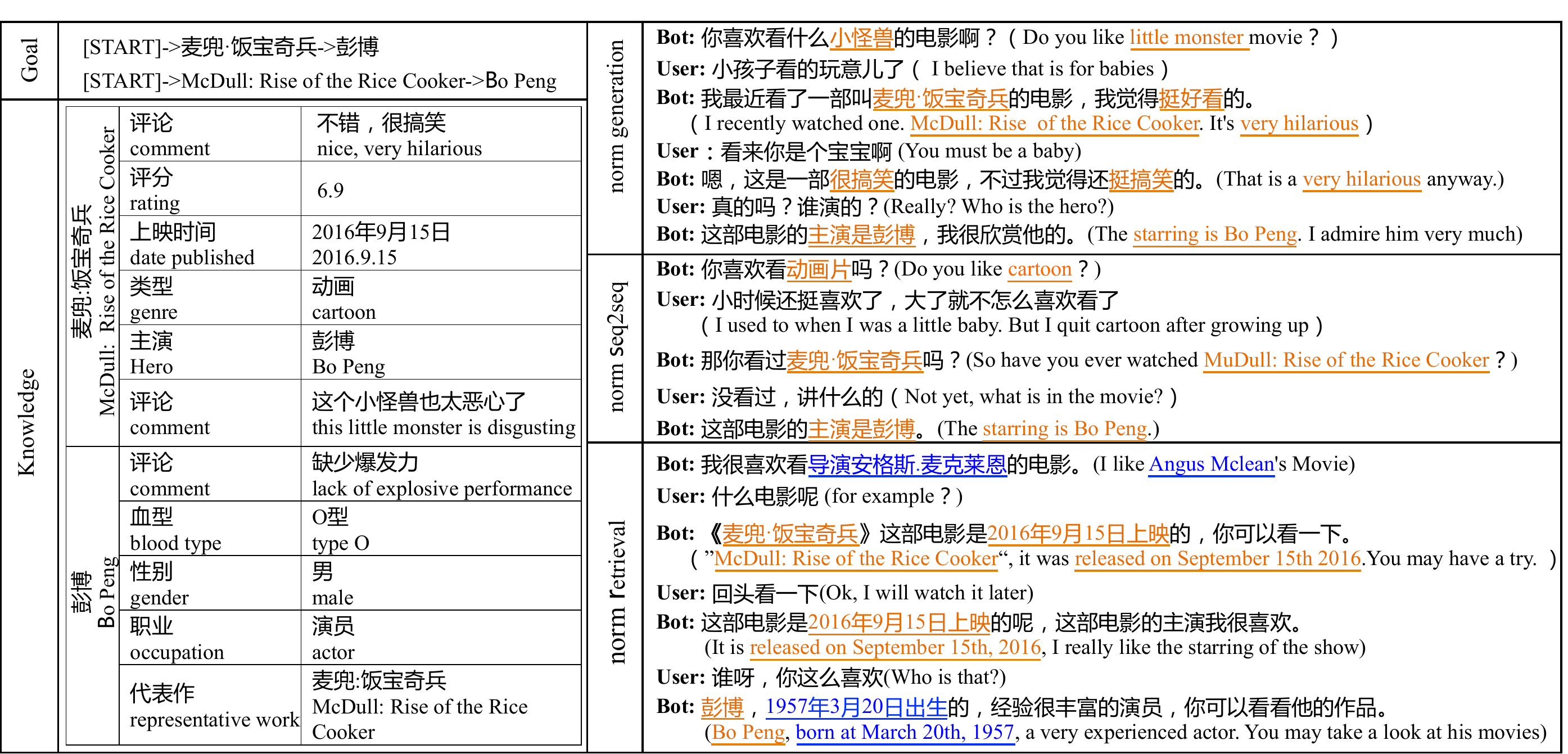}
\caption{Conversations generated by three different models: words in yellow represent correct use of knowledge while those in blue for wrong knowledge. }
\label{fig:analysis-2}
\end{figure*}

From the human evaluation, we found that our proposed generation methods outperform the baseline Seq2Seq model and the retrieval model, especially in terms of turn-level informativeness and proactivity, and dialogue-level goal completion and coherence. In order to further analyze the relationship between informativeness and goal completion, the detailed distribution of goal completion scores and the numbers of used knowledge triplets are shown in Table ~\ref{table:analysis-1}. From this table, it can be seen that our proposed generation model can exploit more knowledge to achieve the conversation goal  (much higher rate on score ``2"), making the conversation more engaging and coherent. This demonstrates the effectiveness of the knowledge posterior/prior distribution learning. Although the baseline Seq2Seq model can also has good goal completion capability, it usually only uses knowledge directly related to the conversation goal in the conversation process (much higher rate over score ``1"), making the conversation usually dull. 

However, for the dialogue-level human evaluation, there are still 15\% to 20\% of conversation goals not achieved. The reason may be that our models (both retrieval and generation)  have no explicit multi-turn policy mechanism to control the whole conversation flow, which is left for future research.

\section{Case Study}

Figure ~\ref{fig:analysis-2} shows the conversations generated by the models via conversing with humans, given the conversation goal and the related knowledge.  It can be seen that our knowledge-aware generator can choose appropriate and more knowledge for diverse conversation generation. Even though the retrieval-based method can also produce knowledge-grounded responses, the used knowledge is often wrong. Although the seq2seq model can smoothly achieve the given knowledge goal, it always generates generic responses using safe dialogue strategy, as the mentioned knowledge is much smaller than our proposed knowledge-aware generator, making the generated conversation less diverse and sometimes dull.

\section{Conclusion}

In this paper, we build a human-like conversational agent by endowing it with the ability of proactively leading the conversation. To achieve this goal, we create a new dataset named DuConv. Each dialog in DuConv is created by two crowdsourced workers, where one acts as the conversation leader and the other acts as the follower. The leader is provided with a knowledge graph and asked to sequentially change the discussed topics following the given conversation goal, and meanwhile, keep the dialogue as natural and engaging as possible. We establish baseline results on DuConv using several state-of-the-art models. Experimental results show that dialogue models that plan over knowledge graph can make more full use of related knowledge to generate more diverse conversations.  Our dataset and proposed models are publicly available, which can be used as benchmarks for future research on constructing knowledge-driven proactive dialogue systems.

\section*{Acknowledgement}

We sincerely thank the PaddlePaddle development team for helping us build the baseline models. We also would like to thank Ying Chen and Na Chen for helping us to collect the dataset through crowdsourcing. This work was supported by the Natural Science Foundation of China (No.61533018).

\bibliography{acl2019}

\begin{thebibliography}{23}
\expandafter\ifx\csname natexlab\endcsname\relax\def\natexlab#1{#1}\fi

\bibitem[{Bordes et~al.(2016)Bordes, Boureau, and Weston}]{bordes2016learning}
Antoine Bordes, Y-Lan Boureau, and Jason Weston. 2016.
\newblock Learning end-to-end goal-oriented dialog.
\newblock \emph{arXiv preprint arXiv:1605.07683}.

\bibitem[{Chung et~al.(2014)Chung, Gulcehre, Cho, and
  Bengio}]{chung2014empirical}
Junyoung Chung, Caglar Gulcehre, KyungHyun Cho, and Yoshua Bengio. 2014.
\newblock Empirical evaluation of gated recurrent neural networks on sequence
  modeling.
\newblock \emph{arXiv preprint arXiv:1412.3555}.

\bibitem[{Devlin et~al.(2018)Devlin, Chang, Lee, and
  Toutanova}]{devlin2018bert}
Jacob Devlin, Ming-Wei Chang, Kenton Lee, and Kristina Toutanova. 2018.
\newblock Bert: Pre-training of deep bidirectional transformers for language
  understanding.
\newblock \emph{arXiv preprint arXiv:1810.04805}.

\bibitem[{Dinan et~al.(2019)Dinan, Roller, Shuster, Fan, Auli, and
  Weston}]{dinan2018wizard}
Emily Dinan, Stephen Roller, Kurt Shuster, Angela Fan, Michael Auli, and Jason
  Weston. 2019.
\newblock Wizard of wikipedia: Knowledge-powered conversational agents.
\newblock In \emph{International Conference on Learning Representations}.

\bibitem[{Fleiss et~al.(1971)}]{fleiss1971measuring}
J.L. Fleiss et~al. 1971.
\newblock {Measuring nominal scale agreement among many raters}.
\newblock \emph{Psychological Bulletin}, 76(5):378--382.

\bibitem[{Ghazvininejad et~al.(2018)Ghazvininejad, Brockett, Chang, Dolan, Gao,
  Yih, and Galley}]{ghazvininejad2018knowledge}
Marjan Ghazvininejad, Chris Brockett, Ming-Wei Chang, Bill Dolan, Jianfeng Gao,
  Wen-tau Yih, and Michel Galley. 2018.
\newblock A knowledge-grounded neural conversation model.
\newblock In \emph{Thirty-Second AAAI Conference on Artificial Intelligence}.

\bibitem[{Li et~al.(2018)Li, Kahou, Schulz, Michalski, Charlin, and
  Pal}]{li2018towards}
Raymond Li, Samira~Ebrahimi Kahou, Hannes Schulz, Vincent Michalski, Laurent
  Charlin, and Chris Pal. 2018.
\newblock Towards deep conversational recommendations.
\newblock In \emph{Advances in Neural Information Processing Systems}, pages
  9748--9758.

\bibitem[{Liu et~al.(2018)Liu, Chen, Ren, Feng, Liu, and Yin}]{P18-1138}
Shuman Liu, Hongshen Chen, Zhaochun Ren, Yang Feng, Qun Liu, and Dawei Yin.
  2018.
\newblock Knowledge diffusion for neural dialogue generation.
\newblock In \emph{Proceedings of the 56th Annual Meeting of the Association
  for Computational Linguistics (Volume 1: Long Papers)}, pages 1489--1498.

\bibitem[{Mo et~al.(2018)Mo, Zhang, Li, Li, and Yang}]{mo2018personalizing}
Kaixiang Mo, Yu~Zhang, Shuangyin Li, Jiajun Li, and Qiang Yang. 2018.
\newblock Personalizing a dialogue system with transfer reinforcement learning.
\newblock In \emph{Thirty-Second AAAI Conference on Artificial Intelligence}.

\bibitem[{Moghe et~al.(2018)Moghe, Arora, Banerjee, and
  Khapra}]{moghe2018towards}
Nikita Moghe, Siddhartha Arora, Suman Banerjee, and Mitesh~M. Khapra. 2018.
\newblock Towards exploiting background knowledge for building conversation
  systems.
\newblock In \emph{Proceedings of the 2018 Conference on Empirical Methods in
  Natural Language Processing}, pages 2322--2332.

\bibitem[{Paszke et~al.(2017)Paszke, Gross, Chintala, Chanan, Yang, DeVito,
  Lin, Desmaison, Antiga, and Lerer}]{paszke2017automatic}
Adam Paszke, Sam Gross, Soumith Chintala, Gregory Chanan, Edward Yang, Zachary
  DeVito, Zeming Lin, Alban Desmaison, Luca Antiga, and Adam Lerer. 2017.
\newblock Automatic differentiation in pytorch.
\newblock In \emph{NIPS-W}.

\bibitem[{Turing(2009)}]{turing2009computing}
Alan~M Turing. 2009.
\newblock Computing machinery and intelligence.
\newblock In \emph{Parsing the Turing Test}, pages 23--65.

\bibitem[{Vougiouklis et~al.(2016)Vougiouklis, Hare, and
  Simperl}]{vougiouklis2016neural}
Pavlos Vougiouklis, Jonathon Hare, and Elena Simperl. 2016.
\newblock A neural network approach for knowledge-driven response generation.
\newblock In \emph{Proceedings of COLING 2016, the 26th International
  Conference on Computational Linguistics: Technical Papers}, pages 3370--3380.

\bibitem[{Wang et~al.(2018)Wang, Liu, Huang, and Nie}]{wang2018learning}
Yansen Wang, Chenyi Liu, Minlie Huang, and Liqiang Nie. 2018.
\newblock Learning to ask questions in open-domain conversational systems with
  typed decoders.
\newblock In \emph{Proceedings of the 56th Annual Meeting of the Association
  for Computational Linguistics (Volume 1: Long Papers)}, pages 2193--2203.

\bibitem[{Wu et~al.(2017)Wu, Wu, Zhou, and Li}]{wu2017sequential}
Yu~Wu, Wei Wu, Ming Zhou, and Zhoujun Li. 2017.
\newblock Sequential match network: A new architecture for multi-turn response
  selection in retrieval-based chatbots.
\newblock In \emph{Proceedings of the 55th Annual Meeting of the Association
  for Computational Linguistics (Volume 1: Long Papers)}, pages 372--381.

\bibitem[{Yao et~al.(2017)Yao, Zhang, Feng, Zhao, and Yan}]{yao2017towards}
Lili Yao, Yaoyuan Zhang, Yansong Feng, Dongyan Zhao, and Rui Yan. 2017.
\newblock Towards implicit content-introducing for generative short-text
  conversation systems.
\newblock In \emph{Proceedings of the 2017 Conference on Empirical Methods in
  Natural Language Processing}, pages 2190--2199.

\bibitem[{Yin et~al.(2015)Yin, Jiang, Lu, Shang, Li, and
  Li}]{journals/corr/YinJLSLL15}
Jun Yin, Xin Jiang, Zhengdong Lu, Lifeng Shang, Hang Li, and Xiaoming Li. 2015.
\newblock Neural generative question answering.
\newblock \emph{CoRR}, abs/1512.01337.

\bibitem[{Young et~al.(2013)Young, Ga{\v{s}}i{\'c}, Thomson, and
  Williams}]{young2013pomdp}
Steve Young, Milica Ga{\v{s}}i{\'c}, Blaise Thomson, and Jason~D Williams.
  2013.
\newblock Pomdp-based statistical spoken dialog systems: A review.
\newblock \emph{Proceedings of the IEEE}, 101(5):1160--1179.

\bibitem[{Zhang et~al.(2018)Zhang, Dinan, Urbanek, Szlam, Kiela, and
  Weston}]{zhang2018personalizing}
Saizheng Zhang, Emily Dinan, Jack Urbanek, Arthur Szlam, Douwe Kiela, and Jason
  Weston. 2018.
\newblock Personalizing dialogue agents: I have a dog, do you have pets too?
\newblock In \emph{Proceedings of the 56th Annual Meeting of the Association
  for Computational Linguistics (Volume 1: Long Papers)}, pages 2204--2213.

\bibitem[{Zhao et~al.(2017)Zhao, Zhao, and Eskenazi}]{zhao2017learning}
Tiancheng Zhao, Ran Zhao, and Maxine Eskenazi. 2017.
\newblock Learning discourse-level diversity for neural dialog models using
  conditional variational autoencoders.
\newblock In \emph{Proceedings of the 55th Annual Meeting of the Association
  for Computational Linguistics (Volume 1: Long Papers)}, pages 654--664,
  Vancouver, Canada.

\bibitem[{Zhou et~al.(2018{\natexlab{a}})Zhou, Young, Huang, Zhao, Xu, and
  Zhu}]{zhou2018commonsense}
Hao Zhou, Tom Young, Minlie Huang, Haizhou Zhao, Jingfang Xu, and Xiaoyan Zhu.
  2018{\natexlab{a}}.
\newblock Commonsense knowledge aware conversation generation with graph
  attention.
\newblock In \emph{Proceedings of the 27th International Joint Conference on
  Artificial Intelligence}, pages 4623--4629.

\bibitem[{Zhou et~al.(2018{\natexlab{b}})Zhou, Li, Dong, Liu, Chen, Zhao, Yu,
  and Wu}]{zhou2018multi}
Xiangyang Zhou, Lu~Li, Daxiang Dong, Yi~Liu, Ying Chen, Wayne~Xin Zhao, Dianhai
  Yu, and Hua Wu. 2018{\natexlab{b}}.
\newblock Multi-turn response selection for chatbots with deep attention
  matching network.
\newblock In \emph{Proceedings of the 56th Annual Meeting of the Association
  for Computational Linguistics (Volume 1: Long Papers)}, volume~1, pages
  1118--1127.

\bibitem[{Zhu et~al.(2017)Zhu, Mo, Zhang, Zhu, Peng, and
  Yang}]{DBLP:journals/corr/abs-1709-04264}
Wenya Zhu, Kaixiang Mo, Yu~Zhang, Zhangbin Zhu, Xuezheng Peng, and Qiang Yang.
  2017.
\newblock Flexible end-to-end dialogue system for knowledge grounded
  conversation.
\newblock \emph{CoRR}.

\end{thebibliography}
\bibliographystyle{acl_natbib}

\clearpage
\section*{Appendix}
\subsection*{A. Turn-level Human Evaluation Guideline}

\textbf{Fluency} measures if the produced response itself is fluent: 
\begin{itemize}
\item score 0 (bad): unfluent and difficult to understand.
\item score 1 (fair): there are some errors in the response text but still can be understood.
\item score 2 (good): fluent and easy to understand. \\
\end{itemize}

\noindent
\textbf{Coherence} measures if the response can respond to the context: 
\begin{itemize}
\item score 0 (bad): not semantically relevant to the context or logically contradictory to the context.
\item score 1 (fair): relevant to the context as a whole, but using some irrelevant knowledge, or not answering questions asked by the users.
\item score 2 (good): otherwise.\\
\end{itemize}

\noindent
\textbf{Informativeness} measures if the model makes full use of knowledge in the response:
 \begin{itemize}
 \item score 0 (bad): no knowledge is mentioned at all.
 \item score 1 (fair): only one triplet is mentioned in the response.
 \item score 2 (good): more than one triplet is mentioned in the response.\\
\end{itemize}

\noindent
\textbf{Proactivity} measures if the model can introduce new knowledge/topics in conversation:
\begin{itemize}
\item score -1 (bad): some new topics are introduced but irrelevant to the context.
\item score 0 (fair): no new topics/knowledge are used.
\item score 1(good): some new topics relevant to the context are introduced.\\
\end{itemize}

\noindent
\subsection*{B. Dialogue-level Human Evaluation Guideline}
\noindent
\textbf{Goal Completion} measures how good the given conversation goal is finished:
\begin{itemize}
\item score 0 (bad): neither ``topic\_a" nor ``topic\_b"is mentioned in the conversation.
\item score 1 (fair): ``topic\_a" or ``topic\_b" is mentioned , but the whole dialogue is very boring and less than 3 different knowledge triplets are used.
\item score 2 (good): both ``topic\_a" or ``topic\_b"  are mentioned and more than 2 different knowledge triplets are used.\\
\end{itemize}

\noindent
\textbf{Coherence} measures the overall fluency of the whole dialogue:
\begin{itemize}
\item score 0 (bad): over 2 responses irrelevant or logically contradictory to the previous context.
\item score 1 (fair): only 2 responses irrelevant or logically contradictory to the previous context.
\item score 2 (good): only 1 response irrelevant or logically contradictory to the previous context.
\item score 3 (perfect): no response irrelevant or logically contradictory to the previous context.
\end{itemize}

\end{document}